\pdfoutput=1

\documentclass[11pt]{article}

\usepackage[]{acl}

\usepackage{times}
\usepackage{latexsym}
\usepackage{graphicx} 
\usepackage{subcaption}
\usepackage{multicol}
\usepackage{hyperref}
\usepackage{lipsum} 
\usepackage{tcolorbox} 
\usepackage{enumitem}

\usepackage[T1]{fontenc}

\usepackage[utf8]{inputenc}

\usepackage{microtype}

\usepackage{tcolorbox}
\tcbuselibrary{breakable}
\usepackage{amsmath}

\usepackage{algorithm}
\usepackage{algpseudocode}

\title{Bi-Chainer: Automated Large Language Models Reasoning \\ with Bidirectional Chaining}

\author{Shuqi Liu$^{1}$ \qquad Bowei He$^{1}$ \qquad Linqi Song$^{1, 2}$\thanks{*Corresponding author} \\
        $^{1}$Department of Computer Science, City University of Hong Kong \\ $^{2}$ Shenzhen Research Institute, City University of Hong Kong \\
        \texttt{\{shuqiliu4-c, boweihe2-c\}@my.cityu.edu.hk}\\
        \texttt{linqi.song@cityu.edu.hk}}

\begin{document}
\maketitle
\begin{abstract}
Large Language Models (LLMs) have shown human-like reasoning abilities but still face challenges in solving complex logical problems. 
Existing unidirectional chaining methods, such as forward chaining and backward chaining, suffer from issues like low prediction accuracy and efficiency. 
To address these, we propose a bidirectional chaining method, Bi-Chainer, which dynamically switches to depth-first reasoning in the opposite reasoning direction when it encounters multiple branching options within the current direction. Thus, the intermediate reasoning results can be utilized as guidance to facilitate the reasoning process. 
We show that Bi-Chainer achieves sizable accuracy boots over unidirectional chaining frameworks on four challenging logical reasoning datasets. Moreover, Bi-Chainer enhances the accuracy of intermediate proof steps and reduces the average number of inference calls, resulting in more efficient and accurate reasoning.
\end{abstract}

\begin{figure*}[htbp]
\centering
\includegraphics[width=1.0\textwidth]{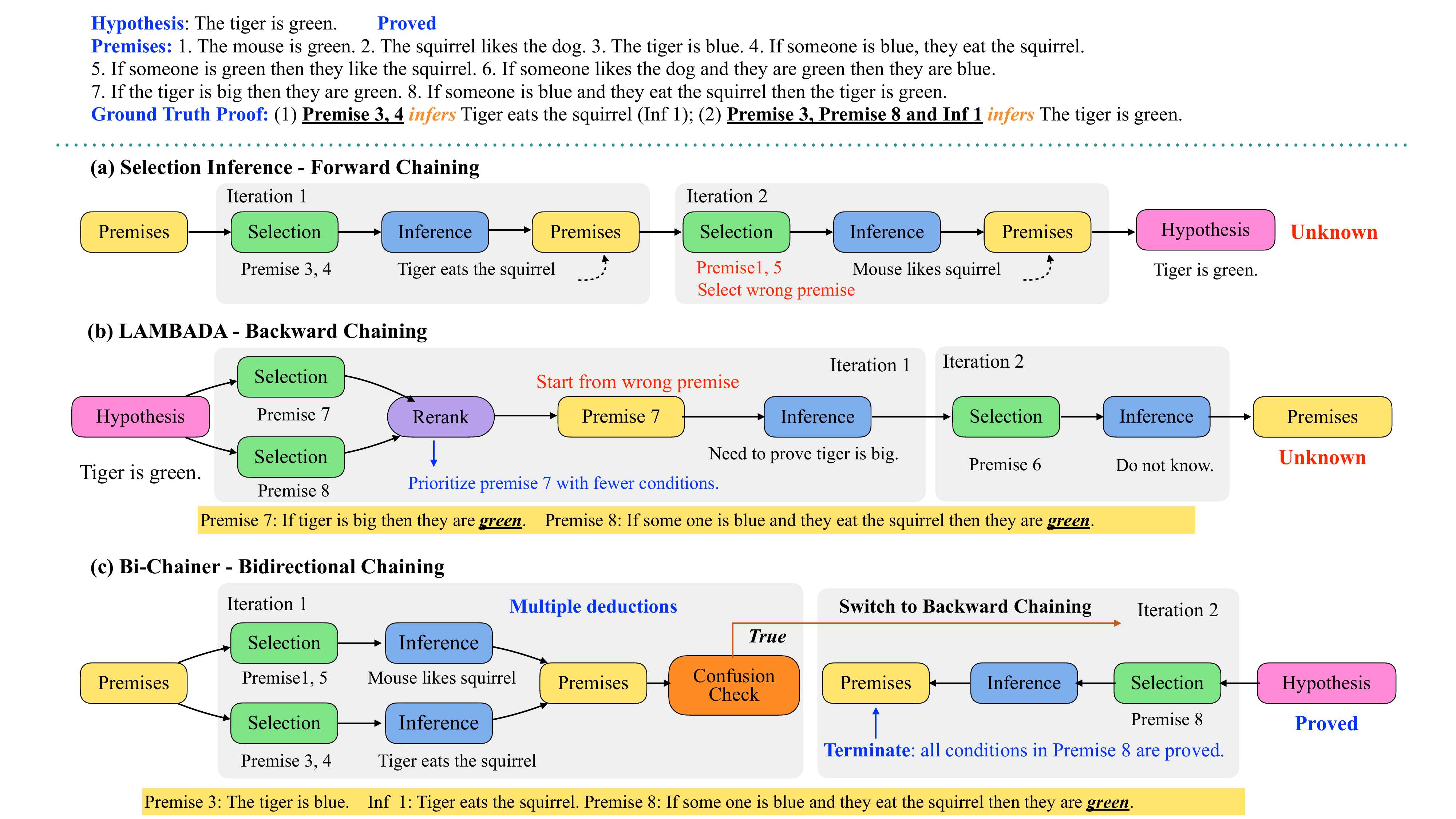}
\caption{Bi-Chainer framework in bidirectional chaining (c) in comparison with the Selection-Inference framework in forward chaining (a) and the LAMBADA framework in backward chaining (b).  }
\label{figure:intro}
\end{figure*}

\section{Introduction}
Automated reasoning involves deriving accurate and valid conclusions from explicitly given knowledge \cite{mccarthy1963programs}. Logical reasoning, particularly in the context of unstructured natural language text, is essential for automated knowledge discovery and has promising implications for advancements in diverse scientific fields. Recently, large language models (LLMs) \cite{touvron2023llama, ouyang2022training, OpenAI2023GPT4TR} have shown promising progress in emulating human-like reasoning abilities \cite{wei2022chain}. However, they still face challenges when it comes to complex multi-step logical reasoning problems \cite{creswell2022selection, kazemi-etal-2023-lambada, valmeekam2022large}.

Recent studies have enhanced reasoning capabilities by employing a modular approach that breaks down complex tasks into smaller, more manageable components. 
Selection-Inference (SI) \cite{creswell2022selection} utilizes forward chaining that employs iterative selection and inference steps to draw conclusions. However, the absence of explicit guidance directly targeting the goal results in subpar and imprecise selection.
On the other hand, LAMBADA \cite{kazemi-etal-2023-lambada} utilizes backward chaining to handle multi-step reasoning. It starts with the goal, recursively selects rules, and iteratively proves decomposed sub-goals. However, LAMBADA's ranking strategy, which prioritizes shorter rules assuming higher success rates, may not always be accurate. This can result in suboptimal performance and hinder the overall efficiency of the reasoning process.

This drives our exploration of the bi-directional chaining method that enables forward chaining with explicit guidance towards the goal and facilitates backward chaining using determinate facts derived from forward chaining. 
As illustrated in Figure \ref{figure:intro}, we present Bi-Chainer, a modular reasoning framework that incorporates bi-directional chaining. 
Bi-Chainer can be understood as a bi-directional depth-first search algorithm that dynamically switches to the opposite reasoning direction when it encounters multiple branching options within the current direction. 
Consequently, the intermediate reasoning outcomes obtained from the opposite reasoning direction can be employed as guidance to enhance the ongoing reasoning process on the current side.

We showcase the adaptability and effectiveness of Bi-Chainer on four logical reasoning datasets: ProofWriter \cite{tafjord2021proofwriter}, FOLIO \cite{han2022folio}, AR-LSAT \cite{zhong2022analytical}, and ParaRules \cite{clark2021transformers}. The datasets encompass a broad range of logical reasoning problems, including deductive reasoning, first-order logic reasoning, and analytical reasoning. 
In addition to achieving quantitative improvements over unidirectional chaining methods, the Bi-Chainer framework also offers qualitative advantages. Firstly, it enhances the accuracy of intermediate proof steps, resulting in more reliable and correct reasoning outcomes at each stage. Secondly, Bi-Chainer reduces the number of inference calls needed during the reasoning process. By utilizing guidance from the opposite side, Bi-Chainer eliminates unnecessary and redundant inference steps.

\section{Related Works}
Recent advancements in large language models, such as LLaMA~\cite{touvron2023llama}, PaLM~\cite{chowdhery2023palm}, and GPT-4~\cite{OpenAI2023GPT4TR}, have demonstrated surprising human-like intelligence in the area of multi-step logical reasoning.
Due to its huge application potential for various applications, including problem-solving, decision-making, and critical thinking~\cite{huang2022towards}, numerous research efforts have been dedicated to improving or eliciting the reasoning ability of these language models. Most of them can be classified into three categories:


\textbf{Fully Supervised Finetuning}: Some previous methods~\cite{rajani2019explain, hendrycks2021measuring} have employed fine-tuning techniques on pre-trained language models (LMs) using downstream datasets to generate rationales or step-by-step solutions, effectively performing reasoning until obtaining final answers. However, these methods heavily rely on meticulously constructed fine-tuning datasets that explicitly capture the reasoning process. Unfortunately, such high-quality data is often difficult to access or requires substantial resources to create. Moreover, this reliance on specific datasets can restrict the extension of the LM's reasoning abilities to other open-ended reasoning tasks beyond the domain of the fine-tuning dataset.

\textbf{Prompting \& In-Context Learning}: The \textit{Chain of Thought (CoT) and its variants}~\cite{wei2022chain} are the most common approaches to release and utilize LLM's reasoning capabilities.
CoT guides the model to generate explicit step-by-step rationale before producing the final results. 
The \textit{rationale engineering} techniques like rational refinement/exploration/verification are complementary to CoT for further eliciting the reasoning capabilities more effectively via refining demonstration rationale examples~\cite{fu2022complexity}, encouraging exploring diverse reasoning ways~\cite{wang2022self}, and verifying if the generated rationales by LLMs lead to correct final answers~\cite{weng2022large}. \textit{Problem decomposition} methods~\cite{zhou2022least, press2022measuring} can also help facilitate the CoT reasoning when tackling complex tasks by decomposing the complex problems into relatively simpler subproblems. It should be mentioned that most previous methods are forward chaining reasoning while only a few works~\cite{kazemi-etal-2023-lambada} have noticed its drawbacks and tried conducting backward chaining reasoning.

\textbf{Hybrid Methods}: Some other methods propose to simultaneously enhance and elicit the reasoning capabilities of LLMs with training and prompting techniques, respectively, like \textit{reasoning-enhanced training and prompting}~\cite{chung2022scaling} and  \textit{bootstrapping \& self-improving}~\cite{zelikman2022star, huang2022large}.

Our work lies in the category of prompting \& in-context learning and aims to fully release the multi-step logical reasoning ability embedded in powerful LLMs like GPT-4. 



\section{Methodology}


In this section, we introduce the Bi-Chainer framework, which automates logical reasoning over natural language premises using bidirectional chaining. The premises $\mathcal{C}$ consist of a set of facts $\mathcal{F}$ and rules $\mathcal{R}$, where rules can be deductive, first-order logic, or analytical reasoning statements. The framework aims to prove or disprove a hypothesis $\mathcal{H}$ based on the given premises. The hypothesis and the premise follow the form ``If $\mathcal{P}$, then $\mathcal{Q}$", where $\mathcal{P}$ represents the condition and $\mathcal{Q}$ represents the consequent.

\subsection{Bi-directional Chaining}

Bidirectional chaining is a reasoning strategy that combines both forward and backward chaining to facilitate the inference process. It involves simultaneous exploration in both directions, starting from the available facts and working forward to derive new conclusions, while also starting from the goal and working backward to decompose the goal into sub-goals using applicable rules.
In our research, we define the existence of multiple deductions or abductions as a confusion state since we aim to ensure a depth-first searching process, thereby reducing the number of LLM calls. In a depth-first search, when faced with multiple deductions or abductions at a single reasoning step, the challenge lies in selecting the most suitable deduction to continue the chaining process. Therefore, we describe this challenge as a confusion in the reasoning process. As the term confusion signifies the need to resolve this ambiguity and make decisions to continue the reasoning chain effectively.

\begin{algorithm}[H]
\caption{Bi-Chainer}
\label{alg:Bi-Chainer}
\begin{algorithmic}[1]
\Require Premises $\mathcal{C} = (\mathcal{F}, \mathcal{R})$, Hypothesis $\mathcal{H}$ with condition $\mathcal{P}$ and consequent $\mathcal{Q}$, Max-Depth D.
\State $\mathcal{F}(\mathcal{H})$ = \textcolor{blue}
{\textit{FactIdentify}}$(\mathcal{H}, \mathcal{F})$

\While {not reach maximum steps D}
\If{ForwardChaining}

   \Repeat

        \State $\mathcal{R}_{d}$ = \textcolor{blue}{\textit{RuleSelection}}($\mathcal{F}(\mathcal{H})$, $\mathcal{R}$, $\mathcal{Q}$)
        \State $\mathcal{F}_{d}$ = \textcolor{blue}{\textit{LogicDeduction}}($\mathcal{F}(\mathcal{H})$, $\mathcal{R}_{d})$
        
        \State Update $\mathcal{F}$ and $\mathcal{F}(\mathcal{H})$ with $\mathcal{F}_{d}$
        \State $v$ = \textcolor{blue}{\textit{FactCheck}}($\mathcal{H}, \mathcal{F})$
        \State $c$ = \textcolor{blue}{\textit{ConfusionCheck}}($\mathcal{F}_{d})$
    \Until{$c$ is True}
    \State Switch to BackwardChaining
\EndIf

\If{BackwardChaining}
\Repeat
    \State $\mathcal{R}_{a}$ = \textcolor{blue}{\textit{RuleSelection}}($\mathcal{Q}$, $\mathcal{R})$
    \State $\mathcal{F}_{a}$ = \textcolor{blue}{\textit{LogicAbduction}}($\mathcal{Q}$, $\mathcal{R}_{a})$
    \State $\mathcal{Q}=\mathcal{F}_{a}$
    \State $v$ = \textcolor{blue}{\textit{FactCheck}}($\mathcal{Q}, \mathcal{F})$
    \State $c$ = \textcolor{blue}{\textit{ConfusionCheck}}($\mathcal{F}_{a})$
\Until{$c$ is True}
\State Switch to ForwardChaining
\EndIf
\If{$v$ is not Unknown}
\State \Return $v$
\EndIf
\EndWhile
\State \Return Unknown
\end{algorithmic}

\end{algorithm}

Figure \ref{figure:intro} illustrates the application of bidirectional chaining in proving a hypothesis using a set of premises. Initially, forward chaining is employed to derive more definite facts and update the premises. In the forward chaining process, deductions are made based on selected premises, such as Premises 3 and 4 leading to the deduction "Tiger eats the squirrel", and Premises 1 and 5 establishing the deduction "Mouse likes squirrel". However, as multiple deductions are obtained, further forward chaining becomes confusing on which deduction to select to continue the chaining process. Therefore, the Confusion Check module triggers a switch to backward chaining.
In the backward chaining phase, both Premise 7 and Premise 8 support the consequence of the hypothesis "Someone is green". However, Premise 8's conditions can all be proven using the intermediate deductions obtained from forward chaining. As a result, the hypothesis is successfully proved using bi-directional chaining.

\subsection{LLM Modules in Bi-Chainer}

To enable applying bidirectioanl chaining for text-based reasoning, we introduce six LLM-based modules: Fact Identification, Rule Selection, Logic Deduction, Logic Abduction, Fact Check, and Confusion Check. Each module is implemented by providing instructions with relevant in-context demonstrations to an LLM (see Appendix \ref{appendix:prompt} for details). We describe these modules and then proceed to the full algorithm.

\noindent{\textbf{Fact Identification Module.}} Given the facts $\mathcal{F}$ from the premises and the hypothesis $\mathcal{H}$, the Fact Identification module is responsible for identifying relevant facts $\mathcal{F}(\mathcal{H}) \in \mathcal{F}$ that contribute to proving the hypothesis.

\noindent{\textbf{Rule Selection Module.}} 
Given a set of rules $\mathcal{R}$ from the premises and a hypothesis $\mathcal{H}$, the Rule Selection module in Forward Chaining identifies a subset of rules $\mathcal{R}_d \in \mathcal{R}$ such that the condition of the rule entails with the facts $\mathcal{F}(\mathcal{H})$ and the consequent of the rule entails with the hypothesis consequent $\mathcal{Q}$. If a rule exists that satisfies these conditions, it is returned as it serves as a bridge between the known facts and the hypothesis, facilitating the concatenation of forward and backward chaining. However, if no such rule is found, only the rules that can be entailed by the known facts are returned. 
The Rule Selection module in Backward Chaining identifies a subset of rules $\mathcal{R}_a \in \mathcal{R}$ such that the consequent of the rule unifies with the consequent of the hypothesis $\mathcal{Q}$. 

\noindent{\textbf{Logic Deduction \& Logic Abduction Modules.}} The Logic Deduction module focuses on deductive reasoning, starting from known facts $\mathcal{F}(\mathcal{H})$ and the deductive rules $\mathcal{R}_d$ to derive a set of new conclusions $\mathcal{F}_d$. 
The Logic Abduction module, on the other hand, deals with abductive reasoning. It aims to generate plausible explanations $\mathcal{F}_a$ that best lead to the hypothesis consequent $\mathcal{Q}$ according to the abductive rules $\mathcal{R}_a$. 
The generated explanations are then treated as new consequences that need to be proven or validated.

\noindent{\textbf{Fact Check.} }
Given the facts $\mathcal{F}$ from the premises, the Fact Check module verifies if the hypothesis $\mathcal{H}$ entails (in which case the hypothesis is proved) or contradicts (in which case the hypothesis is disproved) with the facts. If no such fact can be found, then the truth of $\mathcal{H}$ remains unknown.

\noindent{\textbf{Confusion Check Module.}} The Confusion Check module determines the moment to switch between forward and backward chaining. 
We define a situation where confusion happens when executing the uni-directional chaining at a single step, multiple deductions (in forward chaining) or abductions (in backward chaining) emerge.
In the bidirectional chaining process, if each reasoning step produces consistent deduction $\mathcal{F}_{d}$ or abduction results $\mathcal{F}_{a}$ based on the selected rules, the reasoning continues in that direction. However, if different results emerge at each step, it indicates that the system may be confused in selecting the appropriate rule to proceed with. In such cases, the reasoning is temporarily paused, and the other direction of reasoning is allowed to continue for a few steps to gather additional information that can aid in determining the reasoning path in the current direction.
Bidirectional chaining thus involves continuously switching between forward and backward chaining until a rule is found that connects the consequent of forward chaining with a plausible explanation derived from backward chaining, or until the maximum step limit is reached. 

\subsection{The Bi-Chainer Algorithm}

Algorithm \ref{alg:Bi-Chainer} provides a high-level description of how the six LLM modules described earlier can be integrated with bidirectional chaining to enable text-based logical reasoning (the function calls corresponding to LLM modules are color-coded).

Bi-Chainer can be understood as a bidirectional depth-first algorithm that focuses on reasoning with premises. It employs a depth-first search approach and switches between reasoning directions when faced with multiple branching options.  
Bi-Chainer takes a set of premises $\mathcal{C} = (\mathcal{F}, \mathcal{R})$, a Hypothesis $\mathcal{H}$ with condition $\mathcal{P}$ and consequent $\mathcal{Q}$, and a depth limit $D$ as input. 
The algorithm starts by using the \textit{Fact Identify} module to find facts $\mathcal{F}(\mathcal{P})$ that are essential for proving the hypothesis. It then employs forward chaining to iteratively expand the determinate facts that are associated with and supportive of the hypothesis. 

During Forward Chaining, the Rule Selection module selects rules $\mathcal{R}{d}$ from $R$ that are consistent with the identified facts $\mathcal{F}(\mathcal{H})$. The Logical Deduction module then applies these rules and facts to derive new conclusions $\mathcal{F}{d}$, which are subsequently added to the existing premises. 
The \textit{Fact Check} module then verifies whether the hypothesis can be proved or disproved using the facts. If this is the case, then the algorithm stops and returns the result. 
If not the case, the \textit{Confusion Check} module examines the deduced results to identify any inconsistencies. If different deduction results emerge at each step, it suggests that further deductions based on these conclusions would lead to a significant number of branching paths, deviating from the depth-first approach. In such situations, the algorithm switches the reasoning mode from Forward Chaining to Backward Chaining. 
Similarly, during Backward Chaining, the \textit{Rule Selection} module identifies rules $\mathcal{R}{a}$ from $R$ that unifies with the hypothesis consequent. The \textit{Logical Abduction} module then applies these rules to derive the plausible explanations $\mathcal{F}{a}$, which are then updated as the new consequent to be proved. 
The \textit{Fact Check} module verifies whether the updated hypothesis can be proved or disproved using the facts enriched by Forward Chaining. On the other hand, the \textit{Confusion Check} module examines any inconsistencies are present in $\mathcal{F}{a}$ to determine if a change in the reasoning mode is necessary.

\section{Experimental Setup}

We describe our baselines and datasets here, and provide further implementation details in Appendix B. Unless stated otherwise, all experiments are based on GPT-4 \cite{OpenAI2023GPT4TR}.

\subsection{Baselines}
We compare against the following four baselines.


\textbf{Standard} directly prompts LLM to output labels and proofs in an end-to-end manner, showcasing the lower bound of LLM's capabilities.

\textbf{Chain-of-Thought (CoT)} \cite{wei2022chain}
adopts a step-by-step problem-solving approach, generating explanations before providing the final answer. In our work, the indeterminate explanations are the corresponding step-by-step proof.

\textbf{Selection-Inference (SI)} \cite{creswell2022selection} is a forward modular reasoning framework. 
SI starts from the facts and rules, it iteratively calls selection and inference, until the goal can be proved or disproved.


\textbf{Backward Chaining Reasoning (LAMBADA)} \cite{kazemi-etal-2023-lambada} tackles multi-step reasoning using backward chaining. 
LAMBADA starts from the goal, it recursively selects rules that share the same consequent as the goal and then decomposes the goal into sub-goals based on the antecedent of the selected rules. The recursive selection and decomposition process continues until the sub-goals can be proved or disproved based on the given facts.

\subsection{Datasets}

We experiment with four challenging logical reasoning datasets outlined below.

\textbf{ProofWriter} \cite{tafjord2021proofwriter} is a commonly used synthetic dataset for testing logical reasoning. 
We use the ProofWriter OWA dataset of proof depth 0, 1, 2, 3 and 5.
The task is to determine the provability of the hypothesis as Proved, Disproved, or Unknown based on the given premises. Our reported results include two sets: ProofWriter-PUD, containing all proven examples, and ProofWriter-PD, excluding examples labeled as Unknown.
In line with the methodology outlined by \citet{kazemi-etal-2023-lambada}, we employ the first 1000 examples from the test set for our analysis.

\textbf{FOLIO} \cite{han2022folio} is a challenging expert-written dataset with complex first-order logic reasoning. The problems are mostly aligned with real-world knowledge and use highly natural wordings. 
We use the entire FOLIO test set for evaluation, consisting of 204 examples.


\textbf{AR-LSAT} \cite{zhong2022analytical} is a challenging dataset that focuses on investigating the analytical reasoning of text. The questions are collected from the Law School Admission Test from 1991 to 2016.
We use the entire test set of 230 multiple choice questions. AR-LSAT is particularly challenging, with state-of-the-art models only achieving performance slightly better than random guessing \cite{liang2022holistic, ribeiro2022street}.

\textbf{ParaRules} \cite{clark2021transformers} modifies from ProofWriter where the synthetically generated premises are rewritten by crowdworkers to increase diversity and naturalness.
Thus, we can surpass the evaluation of reasoning limited to templatic expressions. The provided examples necessitate proof depths of up to 5, and the corresponding labels are Proved, Disproved, or Unknown. 
We employ the first 200 examples of the test set for evaluation.


\section{Results}

We now describe the results and compare Bi-Chainer with the baselines in detail.

\subsection{Label Prediction Accuracy}

\begin{figure*}[ht]
  \centering
  \begin{subfigure}[b]{0.5\textwidth}
    \includegraphics[width=\textwidth]{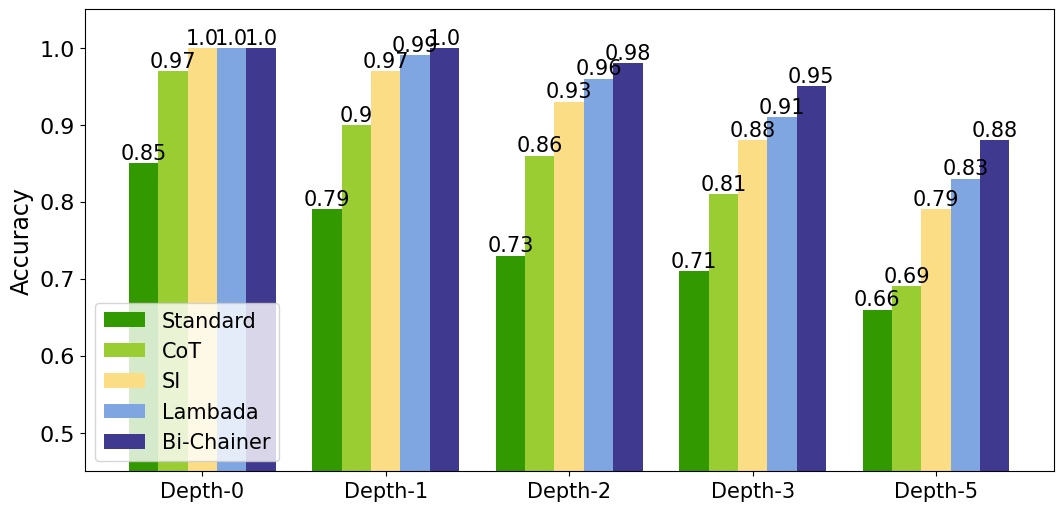}
    \caption{ProofWriter-PUD}
  \end{subfigure}%
  \begin{subfigure}[b]{0.5\textwidth}
    \includegraphics[width=\textwidth]{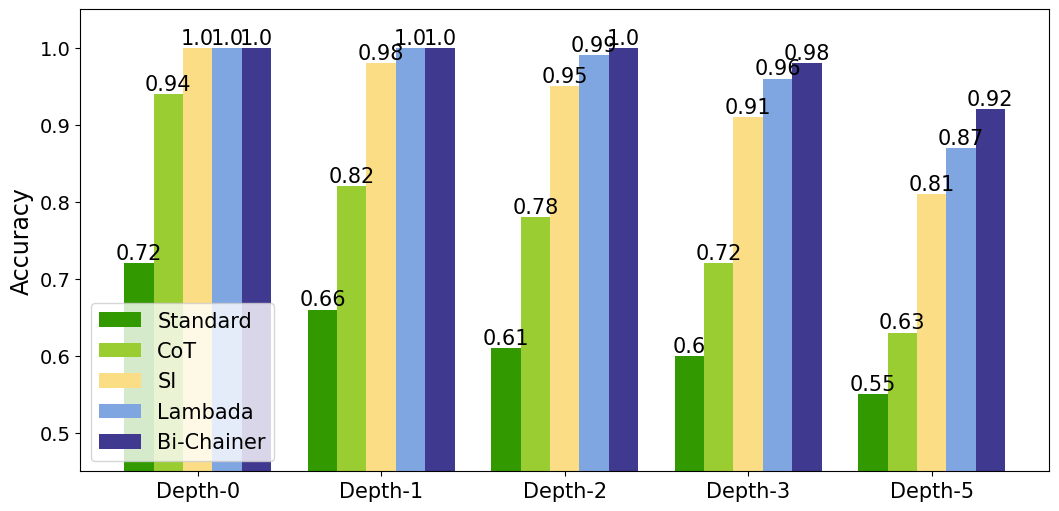}
    \caption{ProofWriter-PD}
  \end{subfigure}

  \begin{subfigure}[b]{0.33\textwidth}
    \centering
    \includegraphics[height=2.8cm]{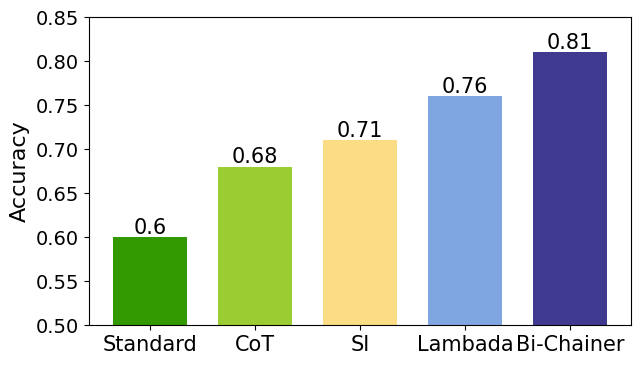}
    \caption{FOLIO}
  \end{subfigure}%
  \begin{subfigure}[b]{0.33\textwidth}
    \centering
    \includegraphics[height=2.8cm]{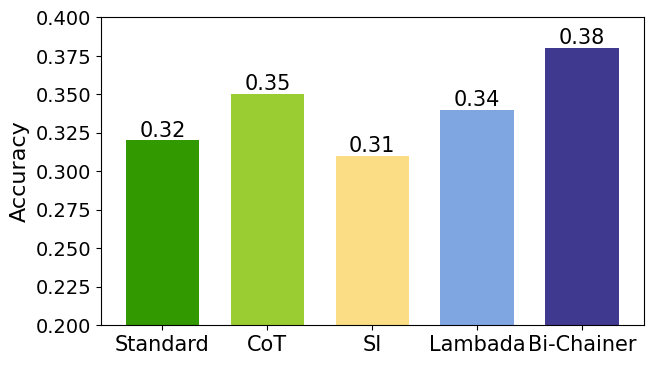}
    \caption{AR-LSAT}
  \end{subfigure}%
  \begin{subfigure}[b]{0.33\textwidth}
    \centering
    \includegraphics[height=2.8cm]{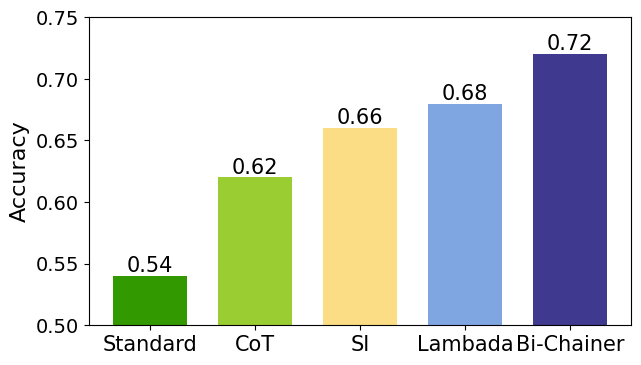}
    \caption{ParaRules}
  \end{subfigure}
  \caption{Label prediction accuracies on (a)-(b) ProofWriter, (c) FOLIO, (d) AR-LSAT, and (e) ParaRules datasets.}
  \label{fig:label_acc}
\end{figure*}

The overall label prediction accuracy results across various reasoning frameworks are reported in Figure \ref{fig:label_acc} (a)-(e). 
The Bi-Chainer framework, employing bi-directional chaining techniques, is observed to significantly outperform both the foundational reasoning models such as the standard and CoT frameworks, and the more advanced,  modular reasoning systems such as the SI and LAMBADA frameworks.
In the evaluation of the ProofWriter-PUD dataset at a reasoning depth of 5, the comparative analysis reveals that Bi-Chainer achieves a relative improvement of 8.9\% over the SI framework. Against the LAMBADA framework, Bi-Chainer maintains a strong lead with a 6.3\% relative improvement. 
Moreover, the FOLIO dataset, which presents more difficult real-world reasoning challenges, also reflects the Bi-Chainer framework's superior performance. Here, Bi-Chainer records a relative improvement of 14.1\% when compared to the SI framework. Against the backward-chaining LAMBADA framework, Bi-Chainer again prevails with a relative improvement of 6.6\%.

In the context of the AR-LSAT dataset, which involves complex analytical reasoning, the modular reasoning frameworks SI and LAMBADA exhibit lower performance compared to CoT. On the other hand, Bi-Chainer demonstrates a relative increase of 8.5\% in performance compared to CoT.
In ParaRules, the introduction of naturalness and diversity through paraphrasing might inadvertently introduce ambiguity of the original premises, resulting in a decrease in the accuracy of label prediction compared to the ProofWriter dataset. However, Bi-Chainer demonstrates a notable relative improvement of 9.1\% over the SI framework and 5.9\% over the LAMBADA framework. 
This consistent outperformance across diverse datasets indicates the adaptability and generalization strength of the Bi-Chainer framework's reasoning mechanisms. 




\begin{figure*}[htb]
  \centering
  \begin{subfigure}[b]{0.48\textwidth}
    \centering
    \includegraphics[height=3.8cm]{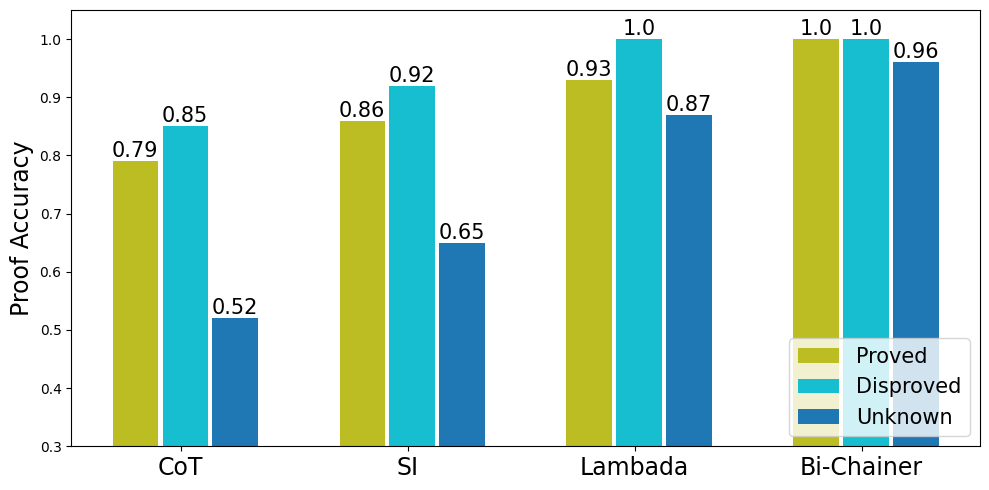}
    \caption{}
    \label{fig:proof_acc_pud}
  \end{subfigure}
  \hfill
  \begin{subfigure}[b]{0.48\textwidth} 
    \centering
    \includegraphics[height=3.8cm]{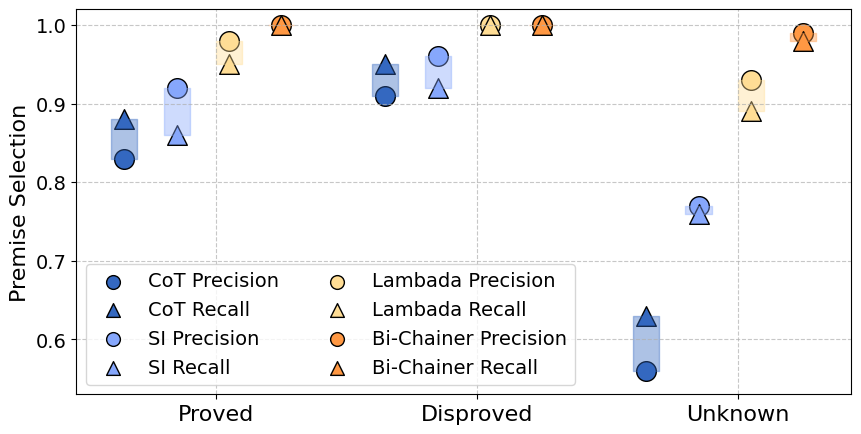}
    \caption{}
    \label{fig:scatter_proof_acc}
  \end{subfigure}
  \caption{
    (a) Proof accuracy results on ProofWriter-PUD (Depth-5) for a set of randomly sampled examples for which the models correctly predicted the goal.
    (b) Precision and Recall results for Premise Selection on the selected samples from the ProofWriter-PUD (Depth-5), with shaded areas indicating the performance gap between different reasoning frameworks for the Proved, Disproved, and Unknown cases.
    }
  \label{fig:proof_acc}
\end{figure*}

\begin{figure*}[h]
\centering

\begin{subfigure}{.24\textwidth}
  \centering
  \includegraphics[height=3.3cm]{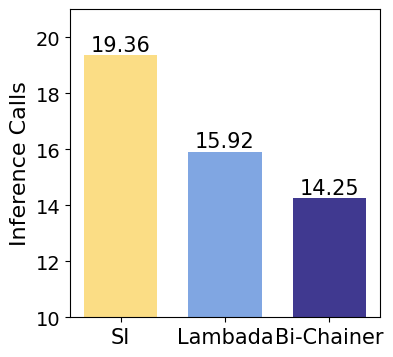}
  \caption{ProofWriter}
  \label{fig:sub1}
\end{subfigure}%
\hfill
\begin{subfigure}{.24\textwidth}
  \centering
  \includegraphics[height=3.3cm]{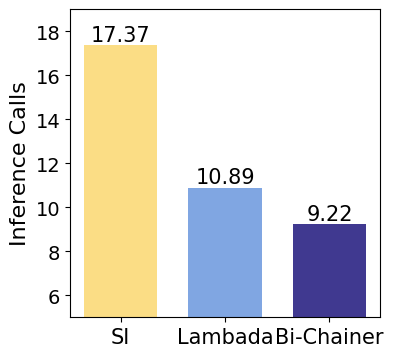}
  \caption{FOLIO}
  \label{fig:sub2}
\end{subfigure}%
\hfill
\begin{subfigure}{.24\textwidth}
  \centering
  \includegraphics[height=3.3cm]{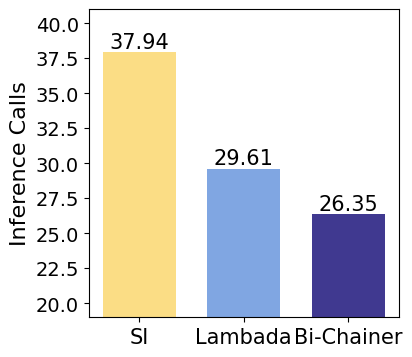}
  \caption{AR-LSAT}
  \label{fig:sub3}
\end{subfigure}
\hfill
\begin{subfigure}{.24\textwidth}
  \centering
  \includegraphics[height=3.3cm]{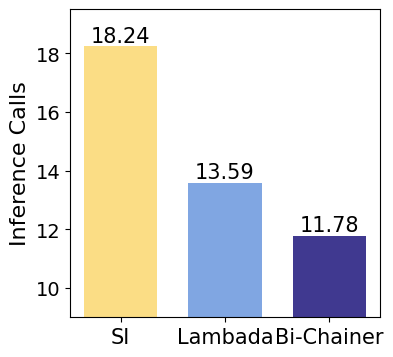}
  \caption{ParaRules}
  \label{fig:sub4}
\end{subfigure}
\caption{Comparing SI, LAMBADA with Bi-Chainer w.r.t. the average number of inference calls they make per example in different datasets.}
\label{fig:visit num}

\end{figure*}

\subsection{Proof Accuracy}

To validate whether each reasoning framework is susceptible to hallucinations, which involve correct final label predictions but incorrect intermediate steps, we conduct an assessment of the proof accuracy. 
We randomly selected separate sets of 50 examples from Depth-5 of the ProofWriter-PUD dataset where each reasoning framework predicted the label correctly and manually verified if the proof chain is correct or not. 
In each step, we compared the facts, rules, and resulting conclusions utilized in the reasoning process to corresponding steps in the reference reasoning path. A proof chain is considered to be correct if these elements are consistent with each other. The proof accuracy results are reported in Figure \ref{fig:proof_acc_pud}.

We observe that different reasoning frameworks demonstrate varying levels of logical reasoning hallucinations in different cases.
In general, modular reasoning frameworks, including SI, Lambada, and Bi-Chainer, are less affected by implicit patterns in language models and achieve higher proof accuracy compared to direct proof generation frameworks like CoT. CoT has an average proof accuracy of 68\%, while SI achieves 78\%. Lambada demonstrates an impressive proof accuracy of 94\%, and Bi-Chainer surpasses all with the highest average proof accuracy of 98\%.
Specifically, we observe that whenever reasoning frameworks predict Proved or Disproved, the prediction is mostly correct. The accuracy is slightly more in cases where the prediction is Disproved. We believe this is because in cases where the result is Disproved, the reasoning path of the model typically involves accurately identifying contradictions or inconsistencies, thereby reducing hallucinatory reasoning.

Moreover, in the case of unknown examples, forward chaining frameworks such as CoT and SI face difficulties in accurately determining the correct reasoning path, achieving relatively low accuracies of 52\% and 65\%, respectively. 
In contrast, the Lambada framework uses backward chaining to capture goal-oriented premises, leading to a significant improvement in achieving 87\% accuracy for unknown cases. On the other hand, the Bi-Chainer framework employs bidirectional chaining to assist premise selection under the guidance of other side's intermediate reasoning results, resulting in an impressive accuracy of 96\%.

In practice, generated reasoning paths often exhibit partial correctness, with errors occurring during the intermediate reasoning process. These errors are mainly attributed to incorrect premise selection, as large models possess powerful single-step reasoning capabilities. To assess the extent of partially correct reasoning, we measure the precision and recall of unique premises extracted from the generated proof that are also present in the reference reasoning path.
The results are presented in Figure \ref{fig:scatter_proof_acc}.
In the case of the CoT method, it heavily relies on internal knowledge and rules within the model to generate proofs, resulting in a limited selection of premises. Consequently, the method exhibits higher recall values (around 5.3\% higher) than precision values. 

On the contrary, the SI method requires considering all available facts and rules that can be used for deduction at each step of the reasoning process. This leads to a larger selection of premises in the reasoning process, resulting in lower recall values (around 3.6\% lower) compared to precision values. 
While most of the reasoning paths in Lambada are correct, in situations where there are numerous and complex facts and rules, there may be a process of error correction. Consequently, the selection of premises in the proof becomes more diverse, leading to lower recall values (around 2.3\% lower) compared to precision values.
In contrast, the Bi-Chainer method excels in handling scenarios with a large number of complex facts and rules. It leverages the guidance provided by the forward chaining process, utilizing intermediate results to guide the backward chaining. This approach mitigates the occurrence of errors and the need for subsequent corrections, resulting in both high recall and precision values.






\subsection{Number of Inference Calls}

Another advantage of Bi-Chainer is its efficiency compared to other modular reasoning frameworks, such as SI and Lambada, which often require multiple LLM inference calls per example. In Figure \ref{fig:visit num}, we compare the average number of LLM calls per example for our datasets.
For the ProofWriter dataset, Bi-Chainer requires 14.25 LLM calls, which is 1.12 times fewer than Lambada and 1.36 times fewer than SI.
In the case of the FOLIO dataset, which has a limited number of premises, Bi-Chainer requires 9.22 LLM calls, exhibiting 1.18 times fewer calls than Lambada and 1.89 times fewer calls than SI.
However, the AR-LSAT dataset poses a different challenge as it contains five options per question. This requires more LLM calls for evaluating each option, resulting in 26.35 calls for Bi-Chainer. Despite this, Bi-Chainer still reduces LLM calls by 1.12 times compared to SI and 1.44 times compared to Lambada.
As for the ParaRules dataset, the presence of paraphrased premises increases the difficulty of accurately selecting the relevant premises. Consequently, the number of LLM calls for ParaRules exhibits a decrease compared to the ProofWriter dataset, with Bi-Chainer requiring 11.78 calls.


\section{Additional Results}

\paragraph{Performance on Open-Source Model.} 
We also adopt the open-sourced LLaMA2 7B model
in greedy search decoding strategy to supplement the corresponding experiments on ProofWriter and ParaRules datasets. 

\begin{table}[htbp]
    \centering
    \small
    \begin{tabular}{ccc}
    \hline
        Method & ProofWriter (d5) & ParaRules \\ \hline
        CoT & 43.4 & 33.5 \\
        SI & 52.6 & 41.0 \\
        Lambada & 58.9 & 43.5 \\
        Bi-Chainer & \textbf{62.3} & \textbf{48.5}\\
        \hline
    \end{tabular}
    \caption{Label accuracy of LLaMA2 7B model on ProofWriter and ParaRules datasets.}
    \label{tab:my_label}
\end{table}

\paragraph{Individual Module Performance.} To understand which components in Bi-Chainer are responsible for the failure cases, we computed the individual accuracy of the six modules described in Section 3. For this purpose, we randomly sampled 100 examples from the validation set of ProofWriter. This sampling included 20 examples for each reasoning depth. We then manually wrote the desired outputs for each module. A module prediction is considered correct if it matches our annotations. The performance of modules in Bi-Chainer is shown in Table \ref{tab:ablation}.

\begin{table}[htbp]
    \centering
    \resizebox{0.48\textwidth}{!}{%
    \begin{tabular}{lcccccc}
    \hline
        Model & FC & FI & RS & LD & LA & CC\\
        \hline
        GPT-4 & 97.82  & 98.78 & 91.52 & 97.44 & 95.26 & 97.73 \\
        LLaMA2  & 83.71 & 86.58 & 65.54 & 93.43 & 89.80 & 95.29  \\
    \hline
    \end{tabular}
    }
    \caption{Individual module performance in Bi-Chainer.}
    \label{tab:ablation}
\end{table}

The evaluation results indicate that the Fact Check module (FC), Fact Identify module (FI), and Confusion Check module (CC) demonstrate a better performance. On the other hand, the Rule Selection module (RS) exhibits the lowest performance among all the modules, indicating that the LLM still faces challenges in effectively selecting the appropriate rules during the reasoning process. Additionally, the Logical Abduction module (LA) performs slightly lower than the Logical Deduction module (LD), suggesting that decomposing conditions are slightly more difficult for the LLM compared to making deduction inferences.

\paragraph{Compare width-first search framework.} Tree of Thoughts (ToT) reasoning framework \cite{yao2024tree} performs reasoning and evaluation on each intermediate result in a tree-searching manner. Thus, compared to our depth-first bi-directional searching framework, ToT is a width-first searching framework, resulting in a high number of inference calls. 
The result comparison between ToT and Bi-Chainer is shown in Table \ref{tab:tot}.

\begin{table}[htbp]
\small
    \centering
    \begin{tabular}{ccc}
    \hline
        Method & Accuracy & Inference calls\\
        \hline
         Standard & 54.0 & 1 \\
         CoT & 61.5 & 1 \\
         \hline
         ToT & 65.0 & 22.79\\
         SI & 65.5 & 18.24\\
         Lambada & 67.5 & 13.59 \\
         Bi-Chainer & \textbf{72.0} & \textbf{11.78} \\
         \hline
    \end{tabular}
    \caption{ToT performance on ParaRules.}
    \label{tab:tot}
\end{table}

The results demonstrate that ToT surpasses SI but trails behind Lambada in terms of performance, and falls even further behind Bi-Chainer. This can be attributed to ToT's focus on solving general complex reasoning tasks, rather than being specifically tailored for goal-oriented tasks like logical reasoning. Besides, ToT's reasoning process, which involves tree-searching for each intermediate result, leads to a significant number of Inference calls.

\paragraph{Robustness Analysis.} We supplement the label accuracy result (mean and standard deviation) of both CoT baseline and our Bi-Chainer under 3 GPT-4 runs on the FOLIO and AR-LSAT datasets in Table \ref{tab:robust}. We observe that the variance across multiple runs is consistently low compared to the improvement in performance, suggesting that GPT-4 is stable in performing logical reasoning tasks.

\begin{table}[h]
    \small
    \centering
    \resizebox{0.45\textwidth}{!}{%
    \begin{tabular}{ccc}
        \hline
        Method & FOLIO & AR-LSAT \\
        \hline
        CoT & 59.64 $\pm$ 0.6112 & 34.93 $\pm$ 1.0974 \\
        Bi-Chainer & \textbf{81.24} $\pm$ 0.8328 & \textbf{38.08} $\pm$ 0.9351 \\
        \hline
    \end{tabular}
    }
    \caption{Label accuracy of CoT and Bi-Chainer on FOLIO and AR-LSAT dataset. We report the mean and standard deviation under 3 GPT-4 runs}
    \label{tab:robust}
\end{table}

\section{Case Study}

We demonstrate a case study to understand the performance of the Bi-Chainer method compared to LAMBADA. We give a high-level overview and abbreviated examples here, leaving full detailed examples in Appendix A.
Lambada experienced premise confusion, it fails to accurately determine the appropriate rule for the subsequent inference step when multiple rules unify with the consequent of the goal statement. As a result of choosing the wrong rule, the model was unable to validate the premise condition, resulting in a wrong conclusion.



\newtcolorbox{lambadacasebox}{
  colback=gray!10,
  colframe=gray!40,
  coltitle=black,
  title=\textbf{Lambada premise confusion case:},
  fonttitle=\bfseries,
  sharp corners,
  boxrule=0.5pt,
  enhanced,
  left=2mm,  
  right=2mm, 
  top=2mm, 
  bottom=2mm
}

\begin{tcolorbox}[breakable]
Hypothesis: The cow chases the cow.



Step 1: 
Rule 2: If someone is rough and the tiger sees the bear then they chase the cow.
Rule 3: If someone likes the tiger then they chase the cow.
\textcolor{blue}{Multiple rules unified.}

Step 2: Select the shorter rule, Rule 2. \textcolor{red}{Select the wrong rule.}


Further steps fail to prove the goal.

Conclusion: \textcolor{red}{Unknown}.

\textbf{Premise confusion error:} Lambada encountered premise confusion where Rule 2 and Rule 3 are both unified with the consequent of the goal statement. \textcolor{red}{The model erroneously selects Rule 2 with fewer sub-goals, leading to further steps that fail to prove the sub-goal.} 


\end{tcolorbox}

\newtcolorbox{biduducorrectcasebox}{
  colback=gray!10,
  colframe=gray!40,
  coltitle=black,
  title=\textbf{Bi-Chainer for Lambada premise confusion},
  fonttitle=\bfseries,
  sharp corners,
  boxrule=0.5pt,
  enhanced,
  left=2mm,  
  right=2mm, 
  top=2mm, 
  bottom=2mm
}

\begin{tcolorbox}[breakable]

\textbf{Bi-Chainer for premise confusion}

Step 1: Identify the facts about the cow, The cow is blue, and The cow chases the lion.



Step 2: In forward-chaining rule selection, we have two candidate rules: Rule 1 and 6.

Step 3: Forward-chaining Logical Deduction: As the cow is blue, we can deduce that the cow chases the tiger from Rule 1. Additionally, since it is stated that the cow chases the lion, we can further deduce that the cows are rough from Rule 6.

\textcolor{blue}{**Detect forward chaining leads to multiple deductions, switch to backward chaining. **}

Step 4: Backward-chaining Rule Selection: we have two candidate rules:
Rule 6: if someone is rough and the tiger sees the bear, then they chase the cow.
Rule 3 states that if someone likes the tiger, then they chase the cow.

Step 4: Backward-chaining Logical Abduction: Using Rule 6 and knowing the cow is rough and the tiger sees the bear, we can deduce that the cow chases the cow. 

Conclusion: \textcolor{blue}{True}. 

\end{tcolorbox}

\section{Conclusion}
We propose the bidirectional chaining method, Bi-Chainer, to overcome the limitations of existing unidirectional chaining methods for complex logical reasoning. By dynamically switching to depth-first reasoning in the opposite direction when faced with multiple branching options, Bi-Chainer leverages intermediate reasoning results to enhance the reasoning process.
In the experiments, Bi-Chainer demonstrates substantial accuracy improvements over unidirectional chaining frameworks on challenging datasets. It also improves the accuracy of intermediate proof steps and reduces the average number of inference calls, resulting in more efficient and accurate reasoning.

\section*{Acknowledgements}

This work was supported in part by the Research Grants Council of the Hong Kong SAR under Grant GRF 11217823 and Collaborative Research Fund C1042-23GF, the National Natural Science Foundation of China under Grant 62371411, InnoHK initiative, the Government of the HKSAR,Laboratory for AI-Powered Financial Technologies.

\section*{Limitations}
This paper presents a novel approach for enhancing reasoning capabilities in large language models through bidirectional chaining. However, it is crucial to acknowledge and address several limitations inherent in this research:

(1) \textbf{Scalability:} The proposed approach may face challenges in terms of scalability when applied to large-scale datasets or real-time applications. The computational complexity of bidirectional chaining may hinder its efficiency, potentially limiting its practicality for scenarios requiring rapid and extensive reasoning.

(2) \textbf{Dependency on Pretrained Models:} The approach heavily relies on pretrained language models, which may introduce certain limitations. Pretrained models are prone to biases and may not capture all relevant contextual information, leading to potential errors or inaccuracies in reasoning outcomes. Additionally, the reliance on pretrained models limits the flexibility and adaptability of the proposed method to new domains or specialized contexts.

(3) \textbf{Lack of Explainability:} While bidirectional chaining enhances reasoning capabilities, it may obscure the interpretability and explainability of the model's decision-making process. Understanding the reasoning steps and how conclusions are reached becomes challenging, hindering transparency and trust in the system. This limitation may impact the acceptance and adoption of the proposed approach in critical applications where interpretability is essential.

(4) \textbf{Knowledge Acquisition and Representation:} The effectiveness of bidirectional chaining heavily depends on the availability and quality of the underlying knowledge base. Incomplete or inaccurate knowledge representations may result in flawed reasoning or incorrect conclusions. Additionally, the challenge of continuously updating and maintaining the knowledge base to keep up with evolving information poses a significant obstacle.

(5) \textbf{Ethical Considerations:} The utilization of large language models raises ethical concerns, including the potential for generating biased or offensive content. Although bidirectional chaining aims to enhance reasoning, it does not inherently address these ethical challenges. Proactive measures, such as comprehensive content filtering and bias detection mechanisms, should be integrated to mitigate the risks associated with unintended outputs.

Addressing these limitations is vital for future research in automated large language models reasoning with bidirectional chaining. Overcoming scalability issues, ensuring model transparency, improving knowledge acquisition, and addressing ethical considerations will contribute to the broader adoption and practicality of the proposed approach in real-world applications.

\section*{Ethics Statement}
This study utilizes publicly available datasets for our models. Prior research endeavors have generally taken ethical considerations into account. We have manually inspected a subset of samples and found no explicit ethical concerns, including violent or offensive content. Nonetheless, it is crucial to highlight that the output generated by large language models lacks the degree of control we might assume. Consequently, we are prepared to implement measures to mitigate any unforeseen outputs.

\bibliography{anthology,custom}
\bibliographystyle{acl_natbib}

\appendix
\newpage
\section{Additional Results and Analyses}
\label{sec:appendix_error}

In this section, we provide some more in-depth qualitative and quantitative analysis of the results from our model and the baselines.

\subsection{Biases of Reasoning Frameworks}

\begin{figure*}[htbp]
    \centering
    \includegraphics[width=1.0\textwidth]{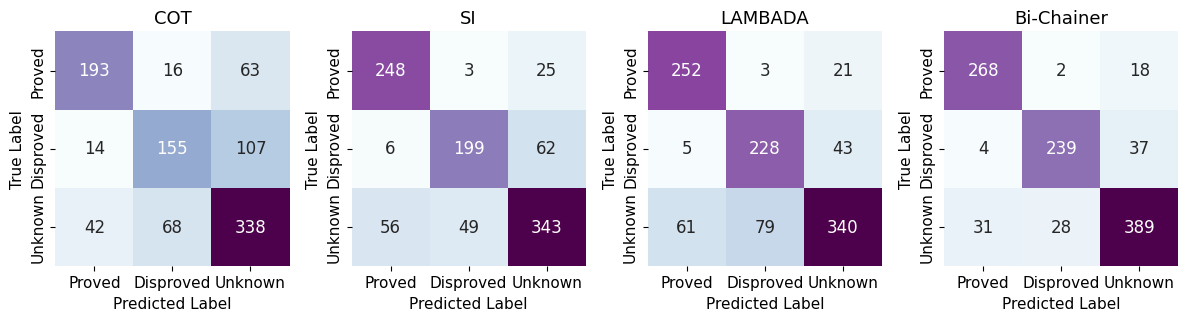}
    \caption{Confusion matrices.}
    \label{figure: confusion mtrix}
\end{figure*}

Figure \ref{fig:label_acc} (a)-(b) demonstrates a performance disparity among different reasoning frameworks when handling datasets with or without Unknown cases. To gain deeper insights into the inherent biases of each method, we provide detailed confusion matrices in Figure \ref{figure: confusion mtrix}. The results reveal that Bi-Chainer consistently outperforms other reasoning frameworks across all Proved, Disproved, and Unknown cases, indicating its ability to achieve accurate and well-balanced predictions. In contrast, CoT exhibits a noticeable bias in predicting Unknown labels, with 24\% of Proved cases and 39\% of Disproved cases being misclassified as Unknown. Consequently, in the absence of unknown cases, the CoT method experiences a decline in model accuracy, while the other methods show an improvement.

Furthermore, we observe that forward chaining is particularly effective in handling Proved cases, while backward chaining demonstrates a more significant improvement in handling Disproved cases. 
Compared to CoT, the Forward chaining-based SI method shows a relative improvement of 29\% for Proved cases and 28\% for Disproved cases. The backward chaining-based Lambada method demonstrates a relative improvement of 31\% for Proved cases and an impressive relative improvement of 47\% for Disproved cases, which is 1.7 times higher than the improvement achieved by the SI method. 
The Bi-Chainer method, which incorporates bidirectional reasoning, combines the advantages of forward chaining that aligns with the natural flow of logical order and backward chaining that focuses on goal-oriented reasoning. It effectively addresses situations of uncertainty in one-directional reasoning by timely incorporating intermediate results from the other side as guidance. This enhances the probability of selecting accurate premises for reasoning. Consequently, the Bi-Chainer method achieves a further relative improvement of 39\% in Proved cases compared to CoT, and a relative improvement of 54\% in Disproved cases.

\subsection{Failure Modes of CoT}
\textbf{CoT demonstrates a self-bias.} CoT tends to rely heavily on its internal knowledge and rules during the reasoning process, disregarding the externally provided facts and rules.

\begin{tcolorbox}[breakable]
\textbf{(GPT-4 Incorrect CoT):} Let's think step by step. From Fact 9 and Rule 4, since the cat sees the bear and someone who sees the cat \textcolor{red}{(which could be anyone, including the cat itself)} sees the bear, they eat the dog. This rule suggests that the cat could eat the dog because the cat sees the bear and \textcolor{red}{the cat sees itself}, fulfilling the condition of the rule. 
Conclusion: Proved

\textbf{Self-bias error:} CoT makes an erroneous inference by concluding that the cat eats the dog based on its own speculation rather than the given facts. The model incorrectly assumes that the cat can see itself, which is not supported by the provided facts. 

\textbf{Hallucinatory Proof:}
However, CoT predict the correct label, indicating self-bias leads to the generation of hallucinatory proofs. 
\end{tcolorbox}

\textbf{CoT fails to make inferences beyond the explicit rules provided. } In complex scenarios where multiple steps of reasoning or indirect relationships are involved, CoT tends to focus on surface-level connections and fails to engage in deep reasoning.

\begin{tcolorbox}[breakable]
Hypothesis: Karen will share "Stranger Things" to Lisa.

Premises:

1. "Stranger Things" is a popular Netflix show.

2. If a Netflix show is popular, Karen will binge-watch it.

3. If and only if Karen binge-watches a Netflix show, she will download it.

4. Karen does not download "Black Mirror".

5. "Black Mirror" is a Netflix show.

6. If Karen binge-watches a Netflix show, she will share it to Lisa.

\textbf{(GPT-4 Incorrect CoT):} Let's think step by step. "Stranger Things" is a popular Netflix show, so Karen will binge-watch it (Premise 1, Premise 2). However, \textcolor{red}{there is no direct information provided about} whether Karen has actually binge-watched "Stranger Things" or whether she has downloaded it. Therefore, we cannot definitively conclude whether Karen will share "Stranger Things" with Lisa based on the given information.

Conclusion: Unknown

\textbf{Multi-step reasoning error:} CoT makes an erroneous inference of the unknown due to its inability to engage in deeper reasoning. CoT gives up after failing to find any direct link between watching "Stranger Things" and sharing with Lisa.
\end{tcolorbox}

\subsection{Failure modes of SI}

\textbf{SI encounters over-inference in the reasoning process.} SI fails to effectively identify and utilize the premise information relevant to validating the hypothesis, leading to excessive processing and reasoning with all available premises, including those that do not contribute to the verification process. This redundancy and unnecessary computation reduce the efficiency and accuracy of SI's reasoning.

\begin{tcolorbox}[breakable]
Hypothesis: The squirrel is blue.

Facts:

1. The dog eats the mouse.\\
2. The dog eats the tiger.\\
3. The dog visits the squirrel.\\
4. The mouse is green.\\
5. The mouse visits the tiger.\\
6. The squirrel is big.\\
7. The squirrel is round.\\
8. The squirrel likes the dog.\\
9. The tiger eats the dog.\\
10. The tiger visits the mouse.\\
Rules:\\
1. If someone eats the tiger and the tiger is big then they are green.\\
2. If someone is green then they like the squirrel.\\
3. If the dog is green then the dog likes the mouse.\\
4. If someone visits the tiger then the tiger is blue.\\
5. If someone visits the tiger then the tiger visits the dog.\\
6. If someone is blue and they eat the squirrel then the squirrel is green.\\
7. If someone is blue then they eat the squirrel.\\
8. If someone likes the dog and they are green then they are blue.\\
\textbf{SI reasoning steps:} \\
Step 1: Analyze Fact 5 and Rule 4. 

Fact 5: the mouse visits the tiger.
Rule 4: if someone visits the tiger then the tiger is blue. 
Deduction: The tiger is blue. \textcolor{blue}{Valid}




Step 2: Analyze the deduced fact that the tiger is blue and Rule 7. \textcolor{blue}{Valid}

Rule 7: if someone is blue, they eat the squirrel.  

Deduction: The tiger eats the squirrel. \textcolor{blue}{Valid}

Step 3: Analyze Fact 4 and Rule 2.  

Fact 4: the mouse is green.
Rule 2: if someone is green then they like the squirrel.

Deduction: Mouse likes the squirrel. \textcolor{red}{Invalid}

Step 4: Analyze Fact 8 and Rule 8.

Fact 8: the squirrel likes the dog.
Rule 8: if someone likes the dog and they are green then they are blue. 

Deduction: \textcolor{red}{We do not have information about the squirrel being green,} so Rule 8 does not apply.  \textcolor{red}{Wrong deduction}

Conclusion: Unknown. 

\textbf{Over-inference error:} Despite deducing that the squirrel is green based on the deduction results in Step 1 and 2 using Rule 6, the presence of invalid reasoning Step 3 prevents the model from progressing along the correct reasoning path within its limited multi-step reasoning capacity. As a result, the model incorrectly concludes that Rule 8 cannot be applied.

\end{tcolorbox}





























\subsection{Failure Modes of LAMBADA}
\textbf{Lambada experienced premise confusion} LAMBADA fails to accurately determine the appropriate rule for the subsequent inference step when multiple rules unify with the consequent of the goal statement. As a result of choosing the wrong rule, the model was unable to validate the premise condition, resulting in a wrong conclusion.
\begin{tcolorbox}[breakable] 
Hypothesis: The cow chases the bear.

Facts:

1. The bear is blue.\\
2. The bear is round.\\
3. The bear sees the cow.\\
4. The cow is blue.\\
5. The lion is rough.\\
6. The lion likes the tiger.\\
7. The lion sees the bear.\\
8. The tiger is cold.\\
9. The tiger is round.\\
10. The tiger sees the bear.\\
11. The tiger sees the cow.\\
Rules:\\
1. If someone is blue then they chase the tiger.\\
2. If the cow is blue and the tiger sees the bear then the cow chases the lion.\\
3. If someone likes the tiger then they chase the lion.\\
4. If someone likes the lion then the lion chases the tiger.\\
5. If the cow is cold and the cow chases the bear then the bear chases the tiger.\\
6. If someone chases the cow and they chase the lion then they chase the bear.\\
7. If someone is rough then they chase the cow.\\
8. If someone is cold then they are blue.\\
9. If someone is blue and they chase the lion then they are rough.\\



\textbf{LAMBADA reasoning steps:}

Step 1: Select Rule 6, If someone chases the cow and they chase the lion then they chase the bear. 

Step 2: We need to prove the cow chases the cow and they chase the lion. 

Step 3: To prove the cow chases the cow, select Rule 7: If someone is rough then they chase the cow.

Step 4: We need to prove the cow is rough.

Step 5: To prove the cow is rough, select Rule 9: If someone is blue and they chase the lion then they are rough.

Step 6: By checking the facts, we know that the cow is blue (Premise 4). 

Step 7: We need to prove the cow chases the lion. 

Step 8: To prove the cow chases the lion, we have two candidate rules. Rule 2: If the cow is blue and the tiger sees the bear then the cow chases the lion; and Rule 3: If someone likes the tiger then they chase the lion.

Step 9: As Rule 3 has fewer sub-goals, we start with proving the cow likes the tiger. \textcolor{red}{Select the wrong rule based on the Rerank strategy in LAMBADA.}

Step 10: Based on the given information, we were \textcolor{red}{unable to find a rule or fact that directly connects or unifies with the statement} "The cow likes the tiger." Therefore, the truth or validity of this statement remains unknown based on the provided context.





Conclusion: Unknown.

\textbf{Premise confusion error:} Lambada encountered premise confusion where Rule 2 and Rule 3 are both unified with the consequent of the goal statement. The model erroneously selects Rule 3 with fewer sub-goals, leading to further steps that fail to prove the sub-goal. 

\end{tcolorbox}

\section{Implementation Details}

For our experiments, we used the GPT-4 \cite{OpenAI2023GPT4TR} for all the models (both Bi-Chainer and the baselines). 
The decoding temperature was set to 0.1.
we limit the maximum number of tokens to generate to 1024 for FOLIO and 4096 for ProofWriter, ParaRules, and AR-LSAT. We use gpt-4-0613 checkpoint of GPT-4 model and invoke the model via the OpenAI API. 
We prompt the model with a set of instructions and 1-8 ICL examples. The examples follow a structured text format designed to scaffold generations and facilitate postprocessing. Each ICL example begins with a task description, followed by the NL hypothesis. The premises are then outlined using numbered statements. The necessary reasoning steps for each example are subsequently outlined in a separate section. 

\noindent{\textbf{ProofWriter}.} We utilize a subset of the publicly available ProofWriter dataset, specifically the Open World Assumption (OWA) dataset \footnote{https://allenai.org/data/proofwriter}.
Due to the cost of inference, we used the first 1000 examples in the test set.
In the Closed World Assumption (CWA) dataset, everything is either proven True or False. However, in the OWA dataset, if a statement cannot be proven True or False, it is labeled as Unknown. For Unknown samples, where there is no explicit reasoning trace, it is essential to enumerate all possible facts. If the hypothesis has not been proven or disproven, it is classified as Unknown.
For this reason, we need to manually verify if the proof chain is correct or not for proof accuracy analysis. 

\noindent{\textbf{FOLIO}.} We use the publicly available FOLIO dataset \footnote{https://github.com/Yale-LILY/FOLIO} and use the validation split of the dataset in our evaluation as the testing split is not publicly available. The original dataset has 204 validation examples.

\noindent{\textbf{AR-LSAT}.} We use the publicly available AR-LSAT dataset \footnote{https://github.com/zhongwanjun/AR-LSAT/tree/main/data} and use the full test set of 230 examples in our evaluation. The AR-LSAT dataset differs from other datasets in that its labels are not fixed. Each example in the AR-LSAT dataset consists of five options associated with a question. To address this, during the prompting process, we concatenate the question with each option to form a hypothesis. Consequently, each AR-LSAT example has five hypotheses that need to be validated. However, the results obtained from validating earlier hypotheses are added to the premises to reduce redundant reasoning among multiple hypotheses.

\noindent{\textbf{Pararules}.} We use a subset of the publicly available ParaRules dataset \footnote{https://allenai.org/data/ruletaker}, specifically the parallel dataset that runs through the Problog reasoner that produced the same labels. The subset consists of the first 200 examples from the test set, with a reasoning depth of 5.  
Table \ref{tab:examples} provides a comprehensive summary of the examples utilized in our study, which are derived from four distinct datasets representing three different types of logical reasoning problems.

\section{Few-Shot Prompts}

We select representative samples from the training split of the dataset as our few-shot examples. These samples are chosen to ensure a balanced representation across different labels. As the training set lacks correct proofs, we manually provide the corresponding proof for each example. We utilize the FOLIO dataset for demonstrating prompts across different reasoning frameworks due to its limited number of premises, which facilitates the presentation.

\subsection{Chain-of-Thought Prompting}

\begin{tcolorbox}[breakable] 
    Task Description:\\
    Given a set of premises, you have to reason whether the hypothesis is true, false, or unknown. 
    \\
    
    Hypothesis:\\
    In La Liga 2021-2022, Real Madrid ranks higher than Barcelona. \\

    Premises:\\
    1: A La Liga soccer team ranks higher than another if it receives more points.\\
    2: If two La Liga soccer teams recieve the same points, the team which recieves more points from the games between the two teams ranks higher.\\
    3: Real Madrid and Barcelona are both La Liga soccer teams.\\
    4: In La Liga 2021-2022, Real Madrid recieves 86 points and Barcelon recieves 73 points.\\
    5: In La Liga 2021-2022, Real Madrid and Barcelona both recieve 3 points from the games between them.\\

    Reason:\\
    Let's think step by step. As indicated by Premise 3, Real Madrid and Barcelona are both La Liga soccer teams.
    From premise 4, Real Madrid received 86 points, and Barcelona received 73 points. This implies Real Madrid has more points than Barcelona.
    From premise 1, if a team receives more points, it ranks higher.
    Therefore, Real Madrid ranks higher than Barcelona based on points. \\

    Answer:\\
    True

\end{tcolorbox}

\subsection{Selection-Inference Prompting}
SI framework iteratively calls the selection and inference module. 
\noindent{The \textbf{selection} prompt is:}
\begin{tcolorbox}[breakable] 
Task Description:\\
Given a set of premises, you have to reason whether the hypothesis is true, false, or unknown. To prove the hypothesis, you need to select the premises where new conclusions can be derived toward proving the goal. \\
    
    Hypothesis:\\
    In La Liga 2021-2022, Real Madrid ranks higher than Barcelona. \\

    Premises:\\
    1: A La Liga soccer team ranks higher than another if it receives more points.\\
    2: If two La Liga soccer teams recieve the same points, the team which recieves more points from the games between the two teams ranks higher.\\
    3: Real Madrid and Barcelona are both La Liga soccer teams.\\
    4: In La Liga 2021-2022, Real Madrid receives 86 points and Barcelon recieves 73 points.\\
    5: In La Liga 2021-2022, Real Madrid and Barcelona both received 3 points from the games between them.\\

    Selected Premises:\\
    Step 1: Premise 3: Real Madrid and Barcelona are both La Liga soccer teams.\\
    Step 2: Premise 4, Real Madrid received 86 points, and Barcelona received 73 points. 
    Step 3: Premise 1, If a team receives more points, it ranks higher.

\end{tcolorbox}

\noindent{The \textbf{inference} prompt is:}

\begin{tcolorbox}[breakable] 
    Task Description:\\
    Derive the inferences based on the selected premises. \\

    Inferences:\\
    Step 1: From Premises 4, Real Madrid received 86 points, and Barcelona received 73 points. This implies Real Madrid has more points than Barcelona.\\
    Step 2: From Premise 1: If a team receives more points, it ranks higher. We know that Real Madrid receives more points than Barcelona. Therefore, Real Madrid ranks higher than Barcelona.
\end{tcolorbox}

\subsection{LAMBADA Prompting}
\label{appendix:prompt}
LAMBADA employs backward chaining with four modules: Fact Check, Rule Selection, Goal Decomposition, and Sign Agreement.
We add instructions for LAMBADA to align with our method, the additional instructions only summarize the main idea of each module.     

The prompt for Fact Chek is:

\begin{tcolorbox}[breakable] 
Task Description:\\
Given a set of premises, you have to reason whether the hypothesis is true, false, or unknown. To prove the hypothesis, you need to check the premises whether the hypothesis can be directly proved or disproved by one of the premises. \\
    
    Hypothesis:\\
    In La Liga 2021-2022, Real Madrid ranks higher than Barcelona. \\

    Premises:\\
    ...\\
    6: In La Liga 2021-2022, Real Madrid ranks higher than Barcelona.\\

    Fact Check:\\
    The hypothesis can be directly proved by Premise 6.
\end{tcolorbox}

The prompt for Rule Selection is:

\begin{tcolorbox}[breakable] 
Task Description:\\
Given a set of premises, you have to reason whether the hypothesis is true, false, or unknown. To prove the hypothesis, you need to select the rules that share the consistent consequences as the hypothesis.  \\
    
    Hypothesis:\\
    In La Liga 2021-2022, Real Madrid ranks higher than Barcelona. \\

    Premises:\\
    1: A La Liga soccer team ranks higher than another if it receives more points.\\
    2: If two La Liga soccer teams recieve the same points, the team which recieves more points from the games between the two teams ranks higher.\\
    3: Real Madrid and Barcelona are both La Liga soccer teams.\\
    4: In La Liga 2021-2022, Real Madrid recieves 86 points and Barcelon recieves 73 points.\\
    5: In La Liga 2021-2022, Real Madrid and Barcelona both recieve 3 points from the games between them.\\

    Rule Selection:\\
    Premise 1, A La Liga soccer team ranks higher than another if it receives more points. or \\
    Premise 2: If two La Liga soccer teams recieve the same points, the team which recieves more points from the games between the two teams ranks higher. 

\end{tcolorbox}

The prompt for Goal Decomposition is:

\begin{tcolorbox}[breakable] 
Task Description:\\
Analyze the plausible sub-goals for the selected rules.  \\
    
    Hypothesis:\\
    In La Liga 2021-2022, Real Madrid ranks higher than Barcelona. \\

    Decomposed Sub-Goals:\\
    According to Premise 1, if we want to prove a La Liga soccer team ranks higher than another, we need to prove the La Liga soccer team receives more points. or \\
    According to Premise 2, if we want to prove a La Liga soccer team ranks higher than another, we need to prove two La Liga soccer teams receive the same points, and one of them receives more points from the games between the two teams. 
\end{tcolorbox}

The prompt for Sign-Agreement is:

\begin{tcolorbox}[breakable] 
Task Description:\\
Check whether the consequence of the rule agrees or disagrees with the hypothesis.  \\
    
    Hypothesis:\\
    In La Liga 2021-2022, Real Madrid ranks higher than Barcelona. \\

    Rule:\\
    In La Liga 2021-2022, Real Madrid ranks higher than Barcelona. \\

    Agreement Sign:\\
    Agree.

\end{tcolorbox}

\subsection{Bi-Chainer Prompting}
Bi-Chainer employs bi-directional chaining with six modules: Fact Check, Fact Identify, Rule Selection, Logical Deduction, Logical Abduction, and Confusion Check.
The prompt for the Fact Check module in our approach aligns with the prompt used in LAMBADA, as presented above.

The prompt for Fact Identify is:

\begin{tcolorbox}[breakable] 
Task Description:\\
Given a set of premises, you have to reason whether the hypothesis is true, false, or unknown. To prove the hypothesis, you need to identify the premises where new conclusions can be derived toward proving the goal. \\
    
    Hypothesis:\\
    In La Liga 2021-2022, Real Madrid ranks higher than Barcelona. \\

    Premises:\\
    1: A La Liga soccer team ranks higher than another if it receives more points.\\
    2: If two La Liga soccer teams recieve the same points, the team which recieves more points from the games between the two teams ranks higher.\\
    3: Real Madrid and Barcelona are both La Liga soccer teams.\\
    4: In La Liga 2021-2022, Real Madrid recieves 86 points and Barcelon recieves 73 points.\\
    5: In La Liga 2021-2022, Real Madrid and Barcelona both recieve 3 points from the games between them.\\

    Fact Identify:\\
    3: Real Madrid and Barcelona are both La Liga soccer teams.\\
    4: In La Liga 2021-2022, Real Madrid recieves 86 points and Barcelon recieves 73 points.\\
    5: In La Liga 2021-2022, Real Madrid and Barcelona both recieve 3 points from the games between them.
    
\end{tcolorbox}

The prompt for Rule Selection in Forward Chaining is:

\begin{tcolorbox}[breakable] 
Task Description:\\
Given a set of premises, you have to reason whether the hypothesis is true, false, or unknown. To prove the hypothesis, you need to select the rules whose conditions entail the identified facts and whose consequents entail the consequent of the hypothesis. If a rule satisfying these criteria is found, return it as the result. Otherwise, return only the rules that are entailed by the identified facts.  \\
    
    Hypothesis:\\
    In La Liga 2021-2022, Real Madrid ranks higher than Barcelona. \\

    Premises:\\
    1: A La Liga soccer team ranks higher than another if it receives more points.\\
    2: If two La Liga soccer teams recieve the same points, the team which recieves more points from the games between the two teams ranks higher.\\
    3: Real Madrid and Barcelona are both La Liga soccer teams.\\
    4: In La Liga 2021-2022, Real Madrid recieves 86 points and Barcelon recieves 73 points.\\
    5: In La Liga 2021-2022, Real Madrid and Barcelona both recieve 3 points from the games between them.\\

    Rule Selection:\\
    Premise 1, A La Liga soccer team ranks higher than another if it receives more points. \\

\end{tcolorbox}

The prompt for Logical Deduction:

\begin{tcolorbox}[breakable] 
    Task Description:\\
    Derive the inferences based on the selected premises. \\

    Inferences:\\
    We know that Real Madrid receives more points than Barcelona (Premise 4). Therefore, Real Madrid ranks higher than Barcelona (Premise 1).

\end{tcolorbox}

The prompt for Rule Selection in Backward Chaining is:

\begin{tcolorbox}[breakable] 
Task Description:\\
Given a set of premises, you have to reason whether the hypothesis is true, false, or unknown. To prove the hypothesis, you need to select the rules whose consequences entail the consequence of the hypothesis.  \\
    
    Hypothesis:\\
    In La Liga 2021-2022, Real Madrid ranks higher than Barcelona. \\

    Premises:\\
    1: A La Liga soccer team ranks higher than another if it receives more points.\\
    2: If two La Liga soccer teams recieve the same points, the team which recieves more points from the games between the two teams ranks higher.\\
    3: Real Madrid and Barcelona are both La Liga soccer teams.\\
    4: In La Liga 2021-2022, Real Madrid recieves 86 points and Barcelon recieves 73 points.\\
    5: In La Liga 2021-2022, Real Madrid and Barcelona both recieve 3 points from the games between them.\\

    Rule Selection:\\
    Premise 1, A La Liga soccer team ranks higher than another if it receives more points. or \\
    Premise 2: If two La Liga soccer teams recieve the same points, the team which recieves more points from the games between the two teams ranks higher.\\

\end{tcolorbox}

The prompt for Logical Abduction:

\begin{tcolorbox}[breakable] 
    Task Description:\\
    Analyze the plausible explanations for the selected rules. \\

    Plausible Reasons:\\
    According to Premise 1, if we want to prove a La Liga soccer team ranks higher than another, we need to prove the La Liga soccer team receives more points. or \\
    According to Premise 2, if we want to prove a La Liga soccer team ranks higher than another, we need to prove two La Liga soccer teams receive the same points, and one of them receives more points from the games between the two teams. 

\end{tcolorbox}

The prompt for Confusion Check:

\begin{tcolorbox}[breakable] 
    Task Description:\\
    Check whether each reasoning step produces consistent deduction or induction results after applying the selected rules. \\

    Abduction Results:\\
    According to Premise 1, if we want to prove a La Liga soccer team ranks higher than another, we need to prove the La Liga soccer team receives more points. or \\
    According to Premise 2, if we want to prove a La Liga soccer team ranks higher than another, we need to prove two La Liga soccer teams receive the same points, and one of them receives more points from the games between the two teams. \\

    Confusion Check: \\ 
    True

\end{tcolorbox}

\begin{table*}[]
\resizebox{\textwidth}{!}{%
\begin{tabular}{cll}
\hline
\multicolumn{1}{l}{Problem} &
  Example &
  Dataset \\
\hline
\begin{tabular}[c]{@{}c@{}}Deductive\\ Reasoning\end{tabular} &
  \begin{tabular}[c]{@{}l@{}}Premises:\\ 1. The bear sees the mouse.\\ 2. The cow visits the dog.\\ 3. The dog visits the cow.\\ 4. The mouse chases the bear.\\ 5. The mouse chases the dog.\\ 6. The mouse is young.\\ 7. The mouse sees the bear.\\ 8. If the mouse is rough and the mouse sees the cow then the mouse is not round.\\ 9. If someone chases the mouse then they see the mouse.\\ 10. If someone is big then they see the dog.\\ 11. If someone is cold and they do not visit the mouse then the mouse sees the dog.\\ 12. If someone sees the mouse then they are big.\\ 13. If someone is young and they visit the cow then the cow does not visit the dog.\\ 14. If someone sees the dog and the dog visits the cow then the cow sees the mouse.\\ 15. If someone sees the dog then the dog sees the bear.\\ Hypothesis:\\ The bear is not big.\end{tabular} &
  \begin{tabular}[c]{@{}l@{}}ProofWriter\\ ParaRules\end{tabular} \\
\hline
\begin{tabular}[c]{@{}c@{}}First-Order\\ Logic\end{tabular} &
  \begin{tabular}[c]{@{}l@{}}Premises:\\ 1: "Stranger Things" is a popular Netflix show\\ 2: If a Netflix show is popular, Karen will binge-watch it\\ 3: If and only if Karen binge-watches a Netflix show, she will download it\\ 4: Karen does not download "Black Mirror"\\ 5: "Black Mirror" is a Netflix show\\ 6: If Karen binge-watches a Netflix show, she will share it to Lisa\\ Hypothesis:\\ Karen will share "Stranger Things" to Lisa.\end{tabular} &
  FOLIO \\
\hline
\begin{tabular}[c]{@{}c@{}}Analytical\\ Reasoning\end{tabular} &
  \begin{tabular}[c]{@{}l@{}}Premises:\\ 1. The organizer of a reading club will select at least five and at most six works \\ from a group of nine works. \\ 2. The group consists of three French novels, three Russian novels, two French plays, \\ and one Russian play. \\ 3. No more than four French works are selected. \\ 4. At least three but no more than four novels are selected. \\ 5. At least as many French novels as Russian novels are selected. \\ 6. If both French plays are selected, then the Russian play is not selected.\\ Hypothesis 1: No Russian novels are selected.\\ Hypothesis 2: Exactly one French novel is selected.\\ Hypothesis 3: All three plays are selected.\\ Hypothesis 4: All three Russian novels are selected.\\ Hypothesis 5: All five French works are selected.\end{tabular} &
  AR-LSAT\\
\hline
\end{tabular}%
}
\caption{ A summary of the examples we use for the four datasets in our study, representing three different types of logical reasoning problems.}
\label{tab:examples}
\end{table*}

\end{document}